\documentclass{article}

\PassOptionsToPackage{numbers, compress}{natbib}
    \usepackage[final]{neurips_2019}

\usepackage[utf8]{inputenc} 
\usepackage[T1]{fontenc}    
\usepackage{hyperref}       
\usepackage{url}            
\usepackage{booktabs}       
\usepackage{amsfonts}       
\usepackage{nicefrac}       
\usepackage{microtype}      
\usepackage{graphicx}
\usepackage{amsmath}
\usepackage{multicol}
\usepackage{xcolor}
\usepackage{float}

\title{Safety and Robustness in Decision Making:\\Deep Bayesian Recurrent Neural Networks for Somatic Variant Calling in Cancer}

\makeatletter
\newcommand{\printfnsymbol}[1]{%
  \textsuperscript{\@fnsymbol{#1}}%
}
\makeatother

\author{
  Geoffroy Dubourg-Felonneau$^1$, Omar Darwish$^{1,2}$, Christopher Parsons$^1$, D\`{a}mi Rebergen$^1$,\\
  \textbf{John W Cassidy$^1$, Nirmesh Patel$^1$, Harry W Clifford$^1$} \\
  \\
  \and
  $^1$Cambridge Cancer Genomics \\
  Cambridge, UK \\
  www.ccg.ai
  \and
  $^2$The University of Cambridge \\
  Cambridge, UK \\
}

\begin{document}

\maketitle

\begin{abstract}
The genomic profile underlying an individual tumor can be highly informative in the creation of a personalized cancer treatment strategy for a given patient; a practice known as precision oncology. This involves next generation sequencing of a tumor sample and the subsequent identification of genomic aberrations, such as  somatic mutations, to provide potential candidates of targeted therapy. The identification of these aberrations from sequencing noise and germline variant background poses a classic classification-style problem. This has been previously broached with many different supervised machine learning methods, including deep-learning neural networks. However, these neural networks have thus far not been tailored to give any indication of confidence in the mutation call, meaning an oncologist could be targeting a mutation with a low probability of being true. To address this, we present here a deep bayesian recurrent neural network for cancer variant calling, which shows no degradation in performance compared to standard neural networks. This approach enables greater flexibility through different priors to avoid overfitting to a single dataset. We will be incorporating this approach into software for oncologists to obtain safe, robust, and statistically confident somatic mutation calls for precision oncology treatment choices.
\end{abstract}

\section{Introduction}

Cancer is a group of diseases that result from specific changes in the genetic code of otherwise healthy cells. These changes vary from DNA point mutations and insertion/deletion mutations, to copy number aberrations and structural variants. Next generation DNA sequencing (NGS) is now commonly used in clinical practice to identify genetic variants, with oncologists using this information for, among other things, determining the best treatment strategy.

However, variants can be difficult to interpret due to deconvolution of cancer (somatic) mutations from heritable (germline) variants, and sequencing error. As a tumor is the product of an accumulation of mutations over time, mutations can be found in anywhere from 0-100\% of the sequencing reads in a tumor, and overlap with the germline variant frequencies and the error rate from the sequencing process. Machine learning approaches have been previously used to classify somatic mutations from germline or error \cite{Harries2018}. However, neural networks in this space have thus far not been tailored to give any indication of confidence in mutation calls, meaning an oncologist could be targeting a mutation with a low probability of being real. We aim to build upon previous approaches by providing a more informative output, including probability that the classification is correct.

For this, we present the use of Bayesian Neural Networks (BNNs) for somatic variant calling in cancer. Augmenting a bayesian inference approach with the power provided by standard NNs. The goal of a BNN is to learn a posterior probability density function (pdf) over the space $\mathcal{W}$ of the weights. This is contrary to a standard NN, where one tries to make point estimations via maximum likelihood. This usually implies overfitting. Overfitting in a standard NN can be relaxed for example using regularization, that from a probabilistic point of view is equivalent to introducing some prior on the weights. 

Instead of dealing with point estimates, BNNs try to learn entire probability distributions. This implies that with respect to standard NNs we have a measure of uncertainty, making the overall network more robust in its predictions, and giving it the ability to say 'I do not know'. This is an important property for decision making.

In this study, we utilize BNNs to provide oncologists with probability values of mutation calls, to better inform them during treatment selection, without impacting performance.  

\section{Variant Calling}

\subsection{The Task}

NGS produces overlapping short sequences of DNA called reads. For a fixed position in a sequenced genome we can find a number of reads coming either from different cells or duplication of an original DNA strand. We call this number the depth. When, for a given patient, we sequence both tumor and normal tissues for comparison, we expect to see a number of true somatic variants (in the  tumor tissue but not in the normal tissue), although the vast majority of the variations observed are due to sequencing noise. Our aim is to provide an indication of confidence in the differentiation of somatic variants from germline variants and error.

\subsection{The Data}

We process the initial raw data using a bioinformatics pipeline that identifies genomic loci where sequencing data from the tumor significantly differs from the normal. We extract the data in the following form:

\begin{multicols}{2}
        \[
        pair
        =
        \left[
        \begin{array}{ccc|ccc}
            N_{11} & \dots  & N_{1w}  &  T_{11} & \dots  & T_{1w} \\
            \vdots & \ddots & \vdots  &  \vdots & \ddots & \vdots \\
            N_{d1} & \dots  & N_{dw}  &  T_{d1} & \dots  & T_{dw} \\
        \end{array}
        \right]
        \]
        
        $d$ : the depth \\
        $w$ : the width or number of observed loci \\
        $N_{ij}$ : the normal base at position $i$ and depth $d$ \\
        $T_{ij}$ : the tumor base at position $i$ and depth $d$ \\        
    \columnbreak
        
    \begin{figure}[H]
      \centering
      \includegraphics[width=0.15\linewidth]{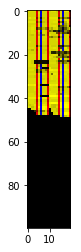}
      \caption{A visual representation of the $pair$ matrix with $w=10$ and $d=100$}
      \label{fig:boat1}
    \end{figure}
        
\end{multicols}

\subsection{The Labels}

For training, we used the Multi-Center Mutation Calling in Multiple Cancers (MC3) high confidence variant dataset \cite{MC3paper}. The labels for this dataset are binary: 1 if the variant is validated positive, 0 if the variant is validated negative. We did not include non-validated data.

\section{Methods and Results}

The input data is the aforementioned $pair$ matrices with a specific size of (100, 20) with three color channels. We reshape the input data from (100, 20, 3) to (100, 60), diluting the color information to analyze using LSTMs. The dataset is also unbalanced, which we correct through undersampling of the majority class. This resulted in a total of 103868 (image, label) pairs. For these data we used a 0.8 training-test split.


We implemented two models: the first including two layers of bidirectional LSTMs and two dense layers, and the second with variational dense layers. We have used DenseFlipout layers provided by the Tensorflow Probability library (tfp)\cite{tfp}\footnote{https://github.com/tensorflow/probability}. Adam optimizer \cite{adamopt} was used to optimize the cost function of each network. 
The usefulness of using variational layers, during training with data $\mathcal{D}$, is to learn an approximate probability density function of the true, usually intractable, posterior $p(\vec{w}|\mathcal{D})$ for the weights of the network. As it is known, the Kullback–Leibler (KL) is loosely speaking a measure of similarities between two probabilty functions. Therefore minimising it, with respect to the parameters $\vec{\theta}$ defining the approximate posterior, would allow us to approximate the true posterior. 

We did this by minimizing a cost function defined as the negative of the ELBO (Evidence Lower Bound) \cite{weightuncertainty}

\begin{equation}
    -ELBO = -\mathbf{E}_{Q_{\vec{w}}(\vec{\theta})}log(P(\mathcal{D}|\vec{w}))+\mathbf{KL}(Q_{\vec{w}}(\vec{\theta})||p(\vec{w}))
    \label{eq:elbo}
\end{equation}{}

where $P(\mathcal{D}|\vec{w})$ and $p(\vec{w})$ are the likelihood of the data and the prior respectively for the weights $\vec{w}$, and $Q_{\vec{w}}(\vec{\theta})$ is an approximate pdf to the true posterior $p(\vec{w}|\mathcal{D})$. Maximizing the ELBO is equivalent to minimizing the KL divergence between the true posterior and the approximate posterior for the weights. We will assume a multivariate gaussian for the variational posterior $Q_{\vec{w}}(\vec{\theta})$. Practically, as the integrals in \ref{eq:elbo} are in general analytically intractable, they are computed via Monte Carlo (MC) sampling from the approximate pdf $Q_{\vec{w}}(\vec{\theta})$. This is automatically done by the variational layers in tfp. Secondly, as it is costly to calculate the ELBO for large datasets such as those in DNA sequencing, tfp allows to do  automatically minibatch optimization. Therefore, \ref{eq:elbo} is calculated as an average over many ELBOs calculated on batches of data $\mathcal{D}_k$ with sampled weights $\vec{w}_i$. This approximation is an unbiased stochastic estimator for the ELBO\cite{weightuncertainty}. As a result, the calculated gradient will not be miscalculated, allowing convergence to the correct solution.

The usefulness of the Flipout estimator \cite{flipout} in the layers is that the unbiased estimation of the ELBO has lower variance with respect to other methods. This is an important point, as the variance of the ELBO dictates the convergence of the optimization process during training. After learning the pdf $Q_{\vec{w}}(\vec{\theta}) \sim p(\vec{w}|\mathcal{D})$ for the weights, via ELBO optimization, we make a new prediction $y^*$ with a new $X^*$ datum, by calculating:


\begin{equation}
    p(y^*) = \int d\vec{w} p(y^*|X^*, \vec{w})p(\vec{w}|\mathcal{D})
\end{equation}{}

This is done in practice by Monte Carlo sampling, where we substitute the integral with an averaged sum over weights sampled from the approximate posterior, to make a prediction from a network in an ensemble of networks, and then taking the average

\begin{equation}
\hat{p}(y^*) = \frac{1}{N_{MC}}\sum_{\vec{w}_{k} \sim  Q_{\vec{w}}(\vec{\theta})} p(y^*|X^*, \vec{w}_k) \label{}
\end{equation}{}

In a classification problem, this means that for each possible outcome $y^* \in \mathcal{A}$, where $\mathcal{A}$ is the set of all outcomes, we will calculate this estimate $\hat{p}(y^*)$, ensuring that $\sum_{y^* \in \mathcal{A}} \hat{p}(y^*) = 1$.

As one of our goals was to check performance degradation, we trained the standard and Bayesian NNs without hyperparameters optimization. We reached a training sparse categorical accuracy of 0.76 and 0.78 for the standard NN and BNN respectively, with a 0.75 accuracy on the test set for both. The standard model does not provide confidence in any given result, while the BNN provides a measure of uncertainty. Utilizing the probabilities for each BNN outcome $y^*$ given the input datum $X^*$, we can generate a 'confidence' measure of the prediction, given by the probability. We show in Figure 2 how the confidence is degraded when a black mask has been artificially added on the input images making the input distribution different than those seen during training. The BNN becomes uncertain for out of distribution samples.


\begin{figure}[H]
    \centering
    \includegraphics[width=15cm]{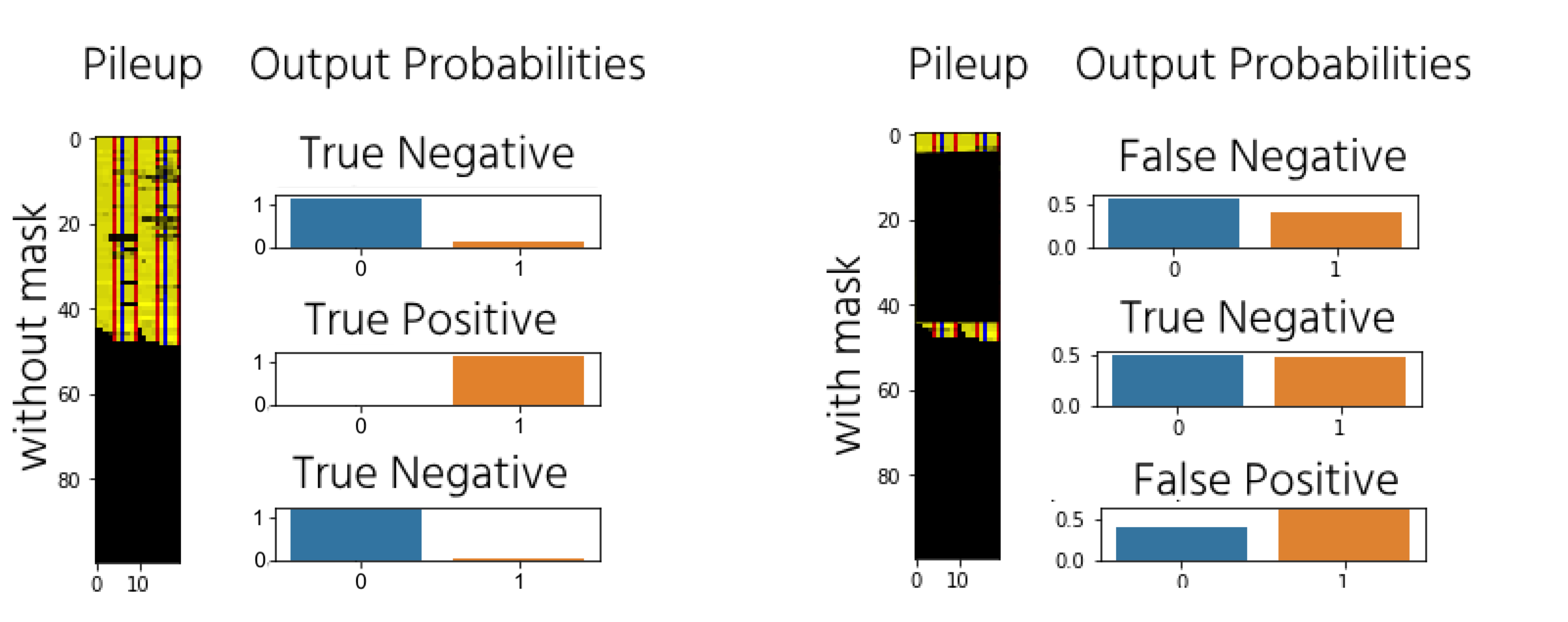} 
    \caption{Sample of results produced by our  Bayesian LSTM applied to original data (left) and masked data (right).}
    \label{fig:bnnfig}
\end{figure}{}

\begin{figure}[H]
    \centering
    \includegraphics[width=15cm]{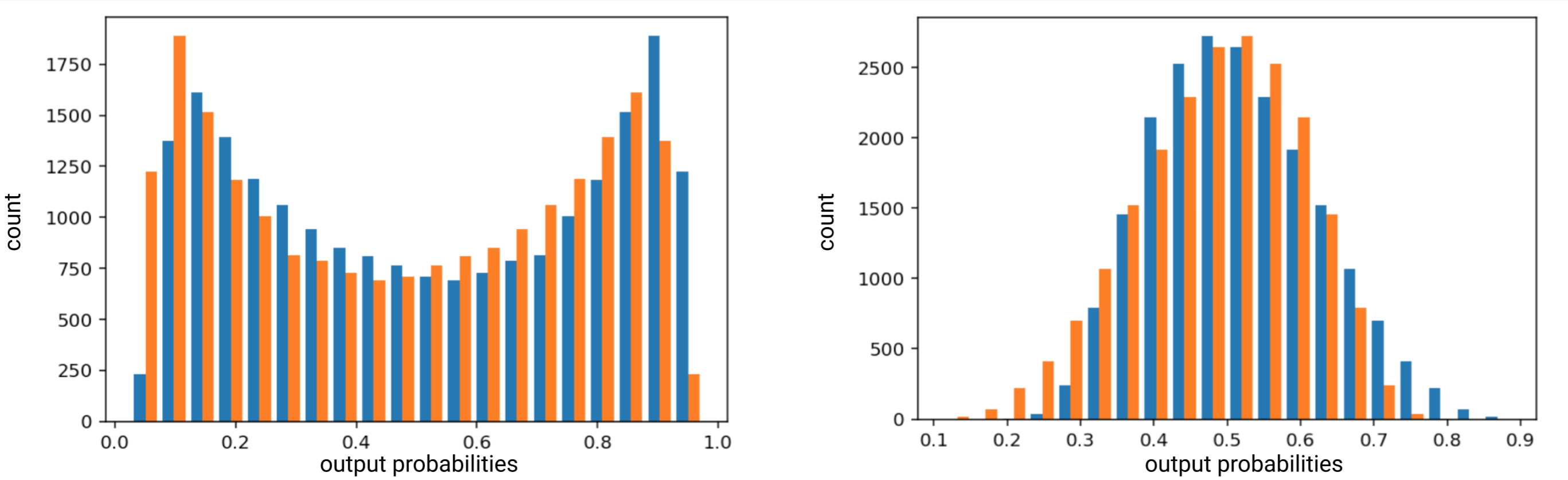} 
    \caption{Histograms of the output probabilities from the BNN tested on in-distribution samples (left) and masked data that are out-of-distribution (right).}
    \label{fig:frequency_test}
\end{figure}

For a better understanding of the BNN behavior we tested our model on both in-distribution and out-of-distribution  samples. The results are summarized in Figure \ref{fig:frequency_test}, where we plot the frequency of the probabilities that the network output for each sample. For the first case, we see clearly a skewed distribution, with a dip in the middle. This means that the BNN is somewhat certain about its predictions. While for the second case, on the right, we observe that output probabilities are mostly around 0.5, showing that the model is uncertain.

{}

\section{Conclusions}

The main results of this paper include: the discovery that BNNs do not degrade the performance of standard NNs in this application, whilst outputting useful probabilities; BNNs enable detection of distribution samples without additional work. Although BNNs are computationally expensive compared to standard NNs, they are more flexible and one can choose different priors or likelihoods to suit the input data, and give naturally regularizing terms (the KL term in \ref{eq:elbo}) to avoid overfitting.

In conclusion, the methods provided herein display a promising novel deep Bayesian neural network methodology for improving genetic variant identification. As a useful approach in providing confidence metrics on neural network derived somatic variant calls, we hope that these methodologies will help clinicians make more informed and better treatment decisions, and eventually show clinical impact in the advancement of precision oncology.

\bibliography{bib}

\end{document}